\documentclass{article}

    \PassOptionsToPackage{numbers, compress}{natbib}
    \usepackage[preprint]{ijcai23}

\usepackage[utf8]{inputenc} % allow utf-8 input
\usepackage[T1]{fontenc}    % use 8-bit T1 fonts
\usepackage{hyperref}       % hyperlinks
\usepackage{url}            % simple URL typesetting
\usepackage{booktabs}       % professional-quality tables
\usepackage{amsfonts}       % blackboard math symbols
\usepackage{nicefrac}       % compact symbols for 1/2, etc.
\usepackage{microtype}      % microtypography
\usepackage{xcolor}         % colors

\usepackage{graphicx}
\usepackage{color}
\usepackage{amsmath}

\newcommand{\fref}[1]{Figure~\ref{#1}}
\newcommand{\tref}[1]{Table~\ref{#1}}

\def\etal{\emph{et al.}}

\title{Brightness-Restricted Adversarial Attack Patch}

\author{%
  Mingzhen Shao \\
  Kahlert School of Computing\\
  University of Utah\\
  Salt Lake City, UT 84108 \\
  \texttt{shao@cs.utah.edu} \\
}

\begin{document}

\maketitle

\begin{abstract}
   % However, the nature of neural networks also exposes it to certain vulnerabilities. 
   % Adversarial attack models have gained increasing attention due to the widespread use of deep neural networks (DNNs), and numerous models have been proposed to deceive these networks, yielding impressive results. 
   % But these models also import many weaknesses due to the nature of deep learning itself. Some adversarial attack models have been proposed to fool the networks and show a good performance. 
   % Among these adversarial attack models, patch-based adversarial attack models stand out because they can be easily utilized in physical world situations and maintain high performance even in complex environmental conditions. 
   % exhibit a high level of robustness across different victim models. 
   % Especially the patch-based adversarial attack models show great robustness in performance across victim models and are easy to deploy in the physical world.  
   % These attack patches usually are bright colors with a moderate size. 
   % However, one of the biggest weaknesses of these models is their conspicuousness.
   Adversarial attack patches have gained increasing attention due to their practical applicability in physical-world scenarios.
   % where they maintain high performance even in complex environmental conditions.
   However, the bright colors used in attack patches represent a significant drawback, as they can be easily identified by human observers. 
   Moreover, even though these attacks have been highly successful in deceiving target networks, which specific features of the attack patch contribute to its success are still unknown. 
   Our paper introduces a brightness-restricted patch (BrPatch) that uses optical characteristics to effectively reduce conspicuousness while preserving image independence. 
   % Our paper introduces a novel approach to generating brightness-restricted patches (BrPatch), which significantly reduces conspicuousness while maintaining image independence.
   % demonstrates that the color lightness of the attack patch is highly redundant. 
   We also conducted an analysis of the impact of various image features (such as color, texture, noise, and size) on the effectiveness of an attack patch in physical-world deployment. 
   % investigate the features of adversarial attack patches used when deploying in the physical world and provide some further methods to reduce the conspicuousness.
   % Additionally, we analyze the features used to deceive victim networks and find that the texture of the patch, rather than its color, plays a critical role.
   % Interestingly, we also observe that the attack patch exhibits significant robustness to random noise, indicating that it may be more difficult to defend against such attacks than previously thought.   
   Our experiments show that attack patches exhibit strong redundancy to brightness and are resistant to color transfer and noise. Based on our findings, we propose some additional methods to further reduce the conspicuousness of BrPatch. Our findings also explain the robustness of attack patches observed in physical-world scenarios.
   % by altering the lightness and color of an attack patch, we can effectively reduce its conspicuousness without requiring any additional finetuning of the attack model. 
   % \footnote{The source code and data are provided in supplementary materials.}
   % Specifically, we found that reducing the lightness of the patch and shifting its color towards a more natural tone can significantly decrease the patch's visibility to human observers, while still maintaining a high success rate in deceiving victim networks.
   % the gradient change in the patches is the key feature that misleads the prediction in victim networks instead of the colors. 
   % Our work sheds light on the mechanisms underlying patch-based adversarial attacks and provides a new avenue for designing more effective attacks that can evade detection.
   % reached. 
\end{abstract}

\section{Introduction}
 Deep neural networks (DNNs) have experienced significant success across various domains in recent years, such as image classsification~\cite{he2016deep, dosovitskiy2020image}, object detection~\cite{he2017mask, redmon2018yolov3}, speech recognition~\cite{wang2017residual}, and natural language processing~\cite{zeng2019dirichlet}. 
 However, the nature of DNNs also makes them vulnerable to adversarial attacks that are crafted by adding carefully designed perturbations on normal examples~\cite{bai2019hilbert, goodfellow2014explaining, ma2018characterizing, szegedy2013intriguing, wang2021convergence, wang2020improving}. 
 Since DNNs have become more critical to some safety-critical applications, such as autonomous driving and biometric authentication, their susceptibility to adversarial attacks raises serious concerns about their safety and reliability~\cite{sharif2016accessorize, evtimov2017robust, dong2019efficient}.

 Adversarial attacks in computer vision tasks can be categorized into two domains: digital and physical. In the digital domain, attackers can access the digital values of inputs and make arbitrary pixel-level changes to inputs. However, such ideal conditions are challenging to achieve in the real world. 
 Data security measures in well-designed software are typically difficult to breach. 
 Once an attacker successfully bypasses these measures, further manipulation of the DNNs may become unnecessary.
 Physical domain attacks, on the other hand, may be more realistic in practice as they assume that only the physical layer objects, such as the environment or objects that the system interacts with, can be manipulated. 
 % Unlike digital attacks, physical attacks can be implemented even if the attacker does not have direct access to the input data. This makes them a viable threat model for DNN-based systems in the real world.
 
 % There are two main types of adversarial attacks in the physical domain: image-dependent and image-independent attacks. 
 Image-dependent and image-independent are the two main types of attacks in the physical domain.
 % Image-dependent attacks require a case-by-case design for different attack. 
 Image-dependent attacks require a design of each attack tailored to the specific target image. 
 In most conditions, these attacks need to replace the target with a modified object.
 For example, to make DNNs misclassify a stop sign, a modified stop sign image needs to be created and substituted for the original one. 
 However, this approach can be highly inefficient and even impossible if the target is not an image. 
 % can be challenging in practice, as it requires a strict requirement for the target object. 
 % This approach can be challenging in practice, as it requires a strict requirements for the target type. \eg, it is almost impossible to replace a banana with a picture.  
 % The image-dependent attacks need to modify the target itself, \eg to make DNNs misclassify a stop sign, a new modified stop sign must be created and replace the orignal one.
 % On the other hand, image-independent attacks do not require modifying the target object. Instead, the attack is performed by introducing additional objects or elements into the scene that can deceive the DNN. These objects are rained to create a physical-world attack without prior knowledge of the other items within the scene. Such attack patches can be put in any environment to launch an attack without modifying the targets.
 % This approach is more flexible and can be applied to a wider range of scenarios. 
 In contrast, image-independent attacks use an additional object (patch) to get rid of this requirement. 
 The patch is trained to create a physical world attack without prior knowledge of the other items within the scene. The patch can be put in any environment to launch an attack without replacing the targets. 
 % Because this approach is robust in practice and minimizes the requirements for the target, many researchers have expanded its potential applications.

 In spite of the great success the attack patches have shown in deceiving target networks, the bright and vivid coloration of these patches can also be a significant drawback. 
 In many real-world scenarios, attack patches must remain low visibility to human observers, particularly in security applications where an attacker may seek to evade visual surveillance. 
 In order to reduce the visibility of attack patches, a straightforward question to ask is whether their vivid coloration is strictly necessary for a successful attack because reducing this feature can significantly decrease the patch's visibility to human observers.

% Therefore, it is critical to find ways to reduce the visibility of attack patches without compromising their effectiveness in deceiving target networks. Our research has shown that altering the lightness and color of the patch can be a promising approach to achieving this goal. By reducing the conspicuousness of the patch, we can enhance its stealth and make it more difficult for human observers to detect, while still maintaining its ability to deceive deep learning models.
 % To  tackle the aforementioned issues, we developed 
 % In this work, we first investigate the redundancy of lightness in an attack patch. And then, we analyze the different roles different feature (color, texture, noise, size) plays in fooling a victim network. And we also test the real-world performance and black box attack performance of the proposed attack patch.
 In this work, we first introduced a brightness-restricted patch (BrPatch) to reduce visibility while maintaining a high attack success rate. 
 By using optical characteristics (brightness) to minimize detectability, the BrPatch can still maintain image independence, which means the BrPatch does not require additional training for specific scenarios.
 We then conducted an analysis of the impact of various image features (such as color, texture, noise, and size) on the effectiveness of an attack patch in a physical-world deployment. 
 Based on our findings, we proposed a hue mapping method to further reduce the visibility of the BrPatch.
 % Specifically, we examined the redundancy of lightness within the patch and investigated the distinct effects that different features play in the deception process. 
 Furthermore, we evaluated the performance of the proposed BrPatch in the physical world to demonstrate that the BrPatch still achieves an attack success rate comparable to the original adversarial attack patch in the real world.
 % Our research has shown that altering the lightness and color of the patch is an efficient approach to reducing the visibility of attack patches without compromising their effectiveness in deceiving victim networks.
 % their efficacy regardless of the surrounding context.
 To the best of our knowledge, this study is the first attempt to use optical characteristics (brightness) to reduce the visibility of an adversarial attack patch and analyze the mechanisms underlying its effectiveness.
 % analyze the mechanisms underlying the effectiveness of an adversarial attack patch.
 % as well as propose a novel approach for reducing its visibility without requiring the generation of new patches for different conditions. 
 
 The main contributions of this paper can be summarized as follows:
 \begin{itemize} 
     \item We propose a brightness-restricted adversarial attack patch (BrPatch) to reduce visibility while preserving image independence. 
     \item Our experiments show that adversarial attack patches exhibit strong redundancy to brightness restrictions.
     \item We demonstrate that color transfer and random noise in the physical world will not significantly affect the performance of attack patches.
     % \item We analyze the influence of different image features on the patch's effectiveness in deceiving a victim network.
     % impact of various image features on the effectiveness of an attack patch in deceiving a victim network. 
 \end{itemize}

% The existence of adversarial attacks reveals not only the intriguing theoretical properties of CNNs, but also raises serious practical concerns about their deployment in security and safety-critical systems. 

% The imperative need to understand the vulnerabilities of CNNs attracts tremendous interest among machine learning, computer vision, and security researchers. 

\section{Related work}
 Adversarial attacks for deep learning were first introduced by szegedy~\etal~\cite{szegedy2013intriguing}. 
 % Since their seminal work, several other attacking models have been proposed~\cite{}. However, a significant limitation of these methods is that they are only applied to some small problems (\eg, MNIST, CIFAR10). 
 % The feasibility of these methods in real-world scenarios is highly questionable. 
 % To accurately replicate the complexity that an attack model would encounter in real-world environments, Kurakin~\etal~\cite{kurakin2016adversarial} first introduced the ImageNet in adversarial attack models.
 Since the publication of their seminal work, numerous researchers have proposed more efficient methods for generating adversarial attacks.
 These adversarial attacks commonly modify each pixel by only a small amount and can be found using a number of optimization strategies such as the Fast Gradient Sign Method (FGSM)~\cite{goodfellow2014explaining}, Projected Gradient Descent (PGD)~\cite{madry2017towards}, and Skip Gradient (SGD)~\cite{wu2020skip}. Some other attacks seek to modify a small number of pixels in the image~\cite{papernot2016limitations}, or a small patch at a fixed location of the image~\cite{sharif2016accessorize}.
 However, all these approaches assume that the attacker has digital-level access to the inputs, which limits the range of scenarios in which the attacks can be used.

 % The most significant drawback of these methods is the requirement to design a new attacker for each target and replace the target with the materialized attacker. 
 % It is an extremely resource-intensive process, and in some cases, nearly impossible (for example, replacing a banana on a desk with a printed picture).
 
 Therefore, Kurakin~\etal~\cite{kurakin2018adversarial} proposed the first physical-domain attack model by printing digital adversarial examples onto paper. They found that a significant portion of the printed adversarial examples deceived the image classifier. 
 Athalye~\etal~\cite{athalye2018synthesizing} improved on this work by creating adversarial objects that remain effective even when viewed from different angles. They achieved this by modeling small-scale transformations synthetically when generating adversarial perturbations. 
 They demonstrated that the resulting adversarial objects deceived target classifiers. In their paper, they claimed that the algorithm is robust to rotations, translations, and noise as long as the transformation can be modeled synthetically. 
 Eykholt~\etal~\cite{eykholt_2018_robust} also developed an attack algorithm that can generate physical adversarial examples. In contrast to Athalye~\etal, they modeled image transformations both synthetically and physically. Their target dataset included certain image transformations, such as changes in viewing angle and distance. They synthetically applied other transformations, such as changes in lighting, when generating adversarial examples. Their work suggested that by relying solely on synthetic transformations, subtleties in the physical environment can be overlooked,
 % relying solely on synthetic transformations can overlook subtleties in the physical environment, 
 resulting in a less robust attack.
 However, all these works require the design of each attack to be tailored to the specific target image, which dramatically limits their applicability in various scenarios.
 
 In order to provide an image-independent adversarial attack model, Brown~\etal~\cite{brown2017adversarial} proposed a new approach for creating an adversarial attack patch.
 This patch can be placed anywhere within the field of view of a classifier and launch an attack. Because this patch is scene-independent, it allows attackers to create a physical-world attack without prior knowledge of the lighting conditions, camera angle, type of classifier being attacked, or even the other items within the scene.
 Since then, many studies have focused on improving physically feasible attacks aimed at deceiving classifiers or object detectors, such as traffic signs~\cite{evtimov2017robust}, cloaks~\cite{wu2020making}, or vehicles~\cite{zhang2019camou}.
 However, as Brown~\etal mentioned in their paper, these attack patches are not restricted to imperceptible changes. These patches have striking color, which is very conspicuous to human observers. 
 One approach to reducing the visibility of the attack patch has been proposed by Duan~\etal~\cite{duan2020adversarial}. They modified the patch into some natural styles that appear legitimate to human observers. 
 Their approach can be an effective way to evade detection,  but it requires generating a new patch for each attack scenario. 
 This regeneration is contrary to the main advantage of using attack patches, which is the ability to be trained once and deployed universally, without being dependent on specific images. 
 Furthermore, in their physical-world experiments, they chose to replace the target object with the modified image rather than using the patch with the original target. 
 These experiments simplified the 3D position relationships between the patch and the target and also suffered the drawback that some targets cannot be replaced with printed images.
 
 % In this work, we aim to address the conspicuousness of attack patches at the source, by mitigating the bright and vivid colors that are commonly used in such attacks. The proposed BrPatch reduces the conspicuousness while maintaining an image-independent and high attack success rate.
 
% poisoned image
% perturbations 
\section{Method}
 In this secession, we first provide a method to generate the brightness-restricted attack patch (BrPatch). 
 Then we provide an easy hue mapping method in the RGB color model to further reduce the visibility of attack patches.

\subsection{Generating brightness-restricted patch}
 % Given a test image $x\in \mathbf{R}^{w \times h \times c}$ with class label $y$, a DNN classifier F: $\mathbf{R}^{w \times h \times c} \to \{1,...,k\}$ mapping image pixels to a discrete label set, and a target class $y_{adv} \neq y$. A patch based adversarial attack is to find a patch $p$ for image $x$ by solving the objective function:
 % \begin{equation}
 %     y_{adv} = F(x+p)
 % \end{equation}

 % The traditional strategy for finding a targeted adversarial example is as follows: 
 Given an image $x\in {[0,1]}^{w \times h \times c}$ with class $y$, and an attack patch $p$, let $T$ be a transformation function that can involve location, rotation, and scale. 
% We define $T(p,x)$ as the input image obtained by applying patch $p$ with transformation $T$ to the original image $x$.
 We define $T(p,x)$ as the input image obtained by applying $T$ to patch $p$ to get the transformed patch, and then overlaid onto $x$.
 
 % using $T(p,x)$ denotes the input image produced by applying patch $p$ to the original image $x$ with transformation $T$ (\eg, location, rotation, scale). 
 The adversarial loss for a targeted attack can be formed as follows: 
 \begin{equation}
     L_{adv} = log(softmax(Pr(\hat{y}|T(p,x))))
 \end{equation}
 where the $\hat{y}$ is the target class and $\hat{y} \neq y$, $Pr$ is the prediction of the target model with respect to class $\hat{y}$.

 % In order to provide a limited brightness for the generated attack patch, we provide a brightness loss as follows:
 % The brightness of the generated attack patch $V$ can be eaisly calculated with RGB to HSV conversion formula:
 % \begin{equation}
 %     V = max(R, G, B)/255
 % \end{equation}
 In order to reduce the intensity of vivid colors in an image, it is necessary to manipulate the brightness component in the HSB (Hue, Saturation, Brightness) color model. 
 However, to prevent the need for switching between multiple color models, we introduced a brightness-restricting loss function within the RGB color space: 
 % in order to avoid switching between different color models, we proposed a lightness restriction loss function in the RGB color model:
 \begin{equation}
     L_b = log(1-mse(p, p_r))
 \end{equation}
 % \begin{equation}
 %     L_b = \begin{cases} 
 %     0, & \text{if } p_v\leq p_{vt} \\
 %     \|p_v-p_{vt}\|_{2}, & \text{otherwise}
 %     \end{cases}
 % \end{equation}
 where $mse(p, p_r)$ calculates the mean square error between the patch $p$ and a reference patch $p_r$. 
 In printing systems, white is used to represent the absence of any ink. 
 Therefore, we use an all-white patch as the reference patch $p_r$. 
 
 The final loss is a combination of $L_{adv}$ and $L_b$:
 \begin{equation}
     L = L_{adv} + \lambda L_b
 \end{equation}
 where $\lambda$ is a parameter that adjusts the strength of the brightness-restricting loss.
 % define a $patch \ application \ operator \ $ which first applies the transformations $t$ to the patch $p$, and then applies the transformed patch $p$ to the image $x$ at location $l$ (see figure).  {\color{red} We do NOT apply scaling here, rotation ONLY!}

%  To obtain the trained patch $\hat{p}$, we use a variant of the Expectation over Transformation (EOT) framework of Athalye~\etal~\cite{athalye2018synthesizing}. In particular, the patch is trained to optimize the objective function
%  \begin{equation}
%      \hat{p} = arg max_p \mathbf{E_{x\sim X,t\sim T,l\sim L}}[logPr(\hat{y}|A(p,x,l,t))]
%  \end{equation}
%  where $X$ is a training set of images, $T$ is a distribution over transformations of the patch, and $L$ is a distribution over locations in the image. Note that this expectation is over images, which encourages the trained patch to work regardless of what is in the background. This departs from most prior work on adversarial perturbations in the fact that this perturbation is $universal$ in that it works for any background. Universal perturbations were identified in \cite{moosavi_2017_universal}, but these required changing every pixel in the image, and results were not given in the physical world.

% %%%%%%%%%%%%%%%%%%%%%%%%%%%%%%
% We also consider camouflaged patches which are forced to look like a given starting image. Here we simply add a constraint of the form $||p-p_{orig}||_\infty < \epsilon$ to the patch objective. This will force the final patch to be within $\epsilon$ in the $L_\infty$ norm of some starting patch $p_{orig}$.

\subsection{Hue mapping in RGB color model}
 Given a patch $p$ and a target region on original image $x_t$, calculate the hue difference between the patch and target region:
 \begin{equation}
     \Delta H = ave_c(p) - ave_c(x_t)
 \end{equation}
 where $ave_c(\cdot)$ computes the average value of each channel (red, green, and blue) of the input patch $p$ and the target region $x_t$, respectively. 
 % where $ave_c(\cdot)$ calculates the average value of each channel of the $p$ and $x$, respectively.
 % $p$, $x$ are the hue value of the patch and target image, respectively. And the 
 
 % In order to ensure that the patch matches the target image, we use the $\Delta H$ to adjust the hue value of the patch:
 We use a threshold $H_t$ to prevent potential overflows of values in the patch $p$ after the hue mapping process:
 \begin{equation}
     p' = \begin{cases} 
             p + \Delta H, & \text{if } \Delta H \leq H_t \\
             p + H_t, & \text{otherwise}
     \end{cases}
 \end{equation}
 where $p'$ is the patch after hue mapping. 

\section{Experimental results and analysis}
 In this section, we first outline the experimental setup. 
 Then we show the performance of the proposed BrPatch in different settings and demonstrate that BrPatch can significantly reduce the visibility of attack patches without compromising their effectiveness in deceiving the target network. The experiments also show that adversarial attack patches have strong redundancy to brightness restrictions. 
 After that, we conduct an analysis of the impact of various image features (such as color, texture, noise, and size) on the effectiveness of an attack patch in the physical world deployment. Based on our findings, we use hue mapping to further reduce the visibility. 
 % This section starts by showing how the proposed lightness-restricted patch can significantly reduce the visibility of attack patches without compromising their effectiveness in deceiving victim networks. 
 Finally, we evaluate the robustness of the BrPatch in physical-world attacks. 
 % In this section, we first demonstrate that the proposed shallow color patch can dramatically increase the stealthiness of the patch while maintaining the same success rate. 
 % generate multiple shallow color patches using ResNet50 as the victim structure, each with a different threshold, and then compare the performance between each patch. 
 % we first generate several lightness-restricted patches with different thresholds in ResNet50 and compare the performance change.
 % Then we conduct a comparative analysis of different feature adjustments and showcase the key characteristics of attack patches. 
 % Then we perform a comparative analysis of various feature adjustments and demonstrate some key features of attack patches. 
 % such as color transform, contrast, and texture filtering. Our results show that texture plays a more significant role in the success rate than color.
\subsection{Experimental setup}
 In our experiments, we use a gray-box setting for the threat model: the source and target networks are both ResNet50 networks but were trained separately on ImageNet1K. 
 We perform 40 epochs of training for each setting and select the patch that achieves the best performance as the final output patch.
 The input images have a size of 224x224, and their values range is $[0,1]$. 
 In order to test the attack success rate, we randomly select 1,000 images in ImageNet1K as source images, and apply the attack patch to these images.
 
 For a physical-world test, we use an HP laser printer to print adversarial attack patches and a Google Pixel6a to take photos.
 To further reduce the visibility of the BrPatch during physical deployment, we first capture a photo of the target and apply hue mapping to the BrPatch. Then we print the hue mapped BrPatch \textbf{only} on A4 paper. We place the printed BrPatch next to the target and take another photo that includes both the target and BrPatch.
 % map its hue to the target region before printing. We use regular A4 paper as the carrier for the printed patch. 
 % And a Google Pixel6a smartphone is used to take photos of the target object. 
 All our experiments focus on targeted attacks, as untargeted attacks can be viewed as a special case of targeted attacks where the target can be any valid target value.
 % pretrained ResNet50 \footnote{torchvision.models.resnet50(ResNet50\_Weights.IMAGENET1K\_V1)} 
 % to generate lightness-restricted patches and as the victim network for the white box attack. For the black box attack, we employ both VGG16 \footnote{torchvision.models.vgg16(VGG16\_Weights.IMAGENET1K\_V1)} 
 % and convNeXt \footnote{torchvision.models.convnex\_base(ConvNeXt\_Base\_Weights.IMAGENET1K\_V1)} as the victim networks to test the generalizability of the proposed patch across different network architectures.
 % In the digital domain, the attack image is created by replacing a part of the original image with the proposed patch. We mask the patch to allow it to take any shape, then train over various images, applying a random translation and rotation on the patch in each image, optimizing using gradient descent. 
 % In the physical domain, we use a normal color printer to print the attack patch on A4 paper and put it side by side with the object. 
 % For each training image, we run a $100$ turn, and once the $softmax(Pr(\hat{y}|A(p,x,l,t))) \geq 0.9$, we early stop and go to the next image. 
 % We save an attack patch each time the training data is iterated over during the 20 epochs, and choose the patch with the highest success rate as the final output of the training.
 % The size of the input images is $224 * 224$, and the value of the image is in the range $[0,1]$. 

% \subsection{Performance Across Different Lightness Restrictions}
\subsection{Performance of brightness-restricted patch}

\begin{figure}[h]
    \centering
    \includegraphics[width=0.8\linewidth]{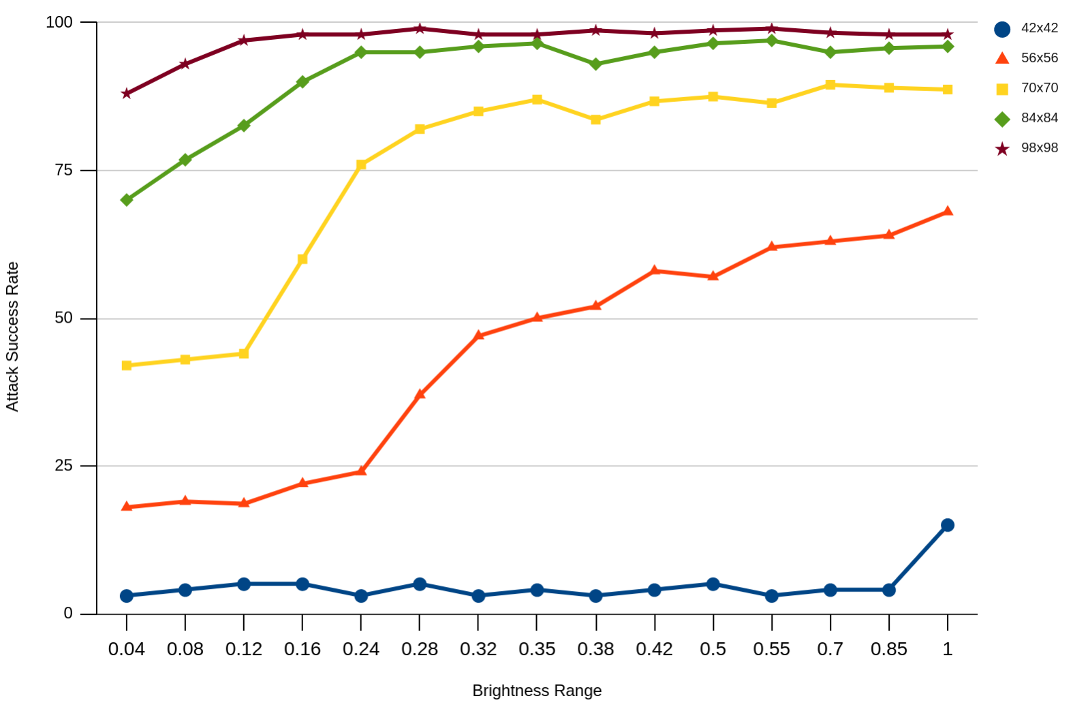}
    \caption{Attack success rate with different patch sizes and brightness ranges.}
    % (b) provides a demonstration for a patch with size 70x70 in different lightness restrictions and corresponding lightness distributions. }
    % Patches and their Lightness Distribution under Varying Lightness Restrictions. 
    % (a)-(c) depict the patches generated with no lightness restriction, a restriction of 0.35, and a restriction of 0.24, respectively. The corresponding lightness distributions are shown in (d)-(f).}
    \label{fig:success_rate}
\end{figure}
 
 \fref{fig:success_rate} shows the performance variations for different brightness ranges and patch sizes on ResNet50, where each line corresponds to a specific patch size. 
 The impact of patch size on the performance is apparent from the figure, with larger patches showing better performance. 
 We also find that all patches show redundancy to brightness, with larger patches being more tolerant to stricter restrictions than smaller ones. 

 % We also observe that all patches show redundancy to brightness restriction, although larger patches are able to tolerate stricter restrictions compared to smaller patches.
 % different patch sizes show varying degrees of sensitivity to the brightness restriction, with larger patches being able to tolerate stricter restrictions compared to smaller patches. 
 % as the lightness restriction becomes stricter, all patches start to exhibit a decline in their success rate.
 % And different patch sizes displayed varying levels of sensitivity to the lightness restriction, with larger patches being able to tolerate a stricter restriction compared to smaller patches.
 \begin{figure}
    \centering
      \includegraphics[width=0.7\linewidth]{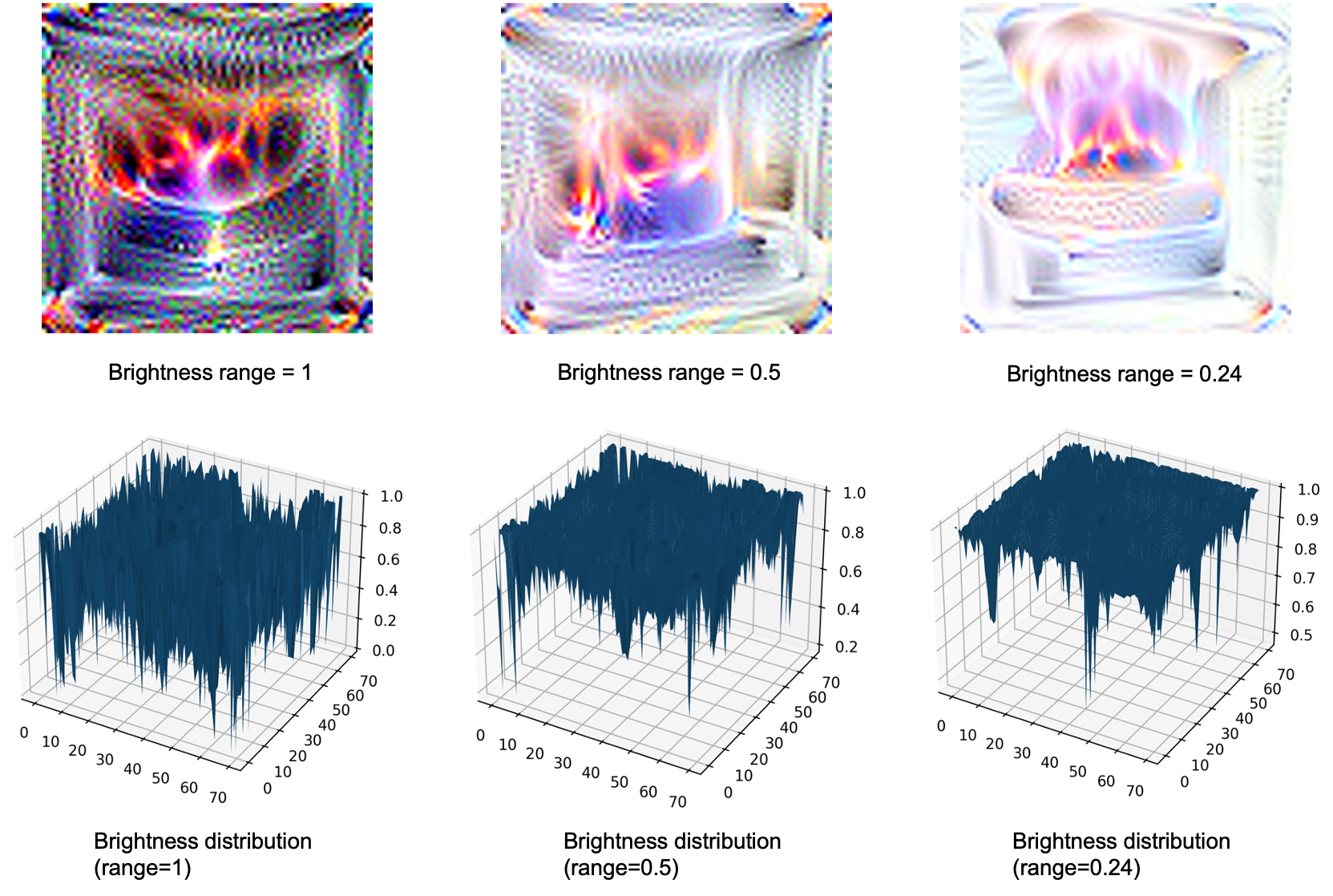}
      \caption{Patches with different brightness restriction and their brightness distribution. (Top: Different brightness-restricted patches; Bottom: Corresponding brightness distributions to each patch.)}
    \label{fig:limit}
\end{figure}

 Some brightness distributions of different BrPatch are shown in \fref{fig:limit}. 
 We choose a patch size of 70x70 due to its high sensitivity to brightness restriction and ability to achieve high performance without brightness restriction (original adversarial attack patch). For the following analysis, we will continue to use attack patches of the same size. 
 
 Referring to the performance shown in \fref{fig:success_rate}, our results indicate that the adversarial patches can achieve a high attack success rate in deceiving the target network with only a small range of brightness values. For instance, the performance remains consistent for the patch sizes of 70x70 even when up to 65\% of the brightness is lost. 
 This finding allows us to decrease the brightness of an attack patch without sacrificing its performance, effectively reducing the visibility of the patch.

\begin{figure}[h]
    \centering
    \includegraphics[width=0.95\linewidth]{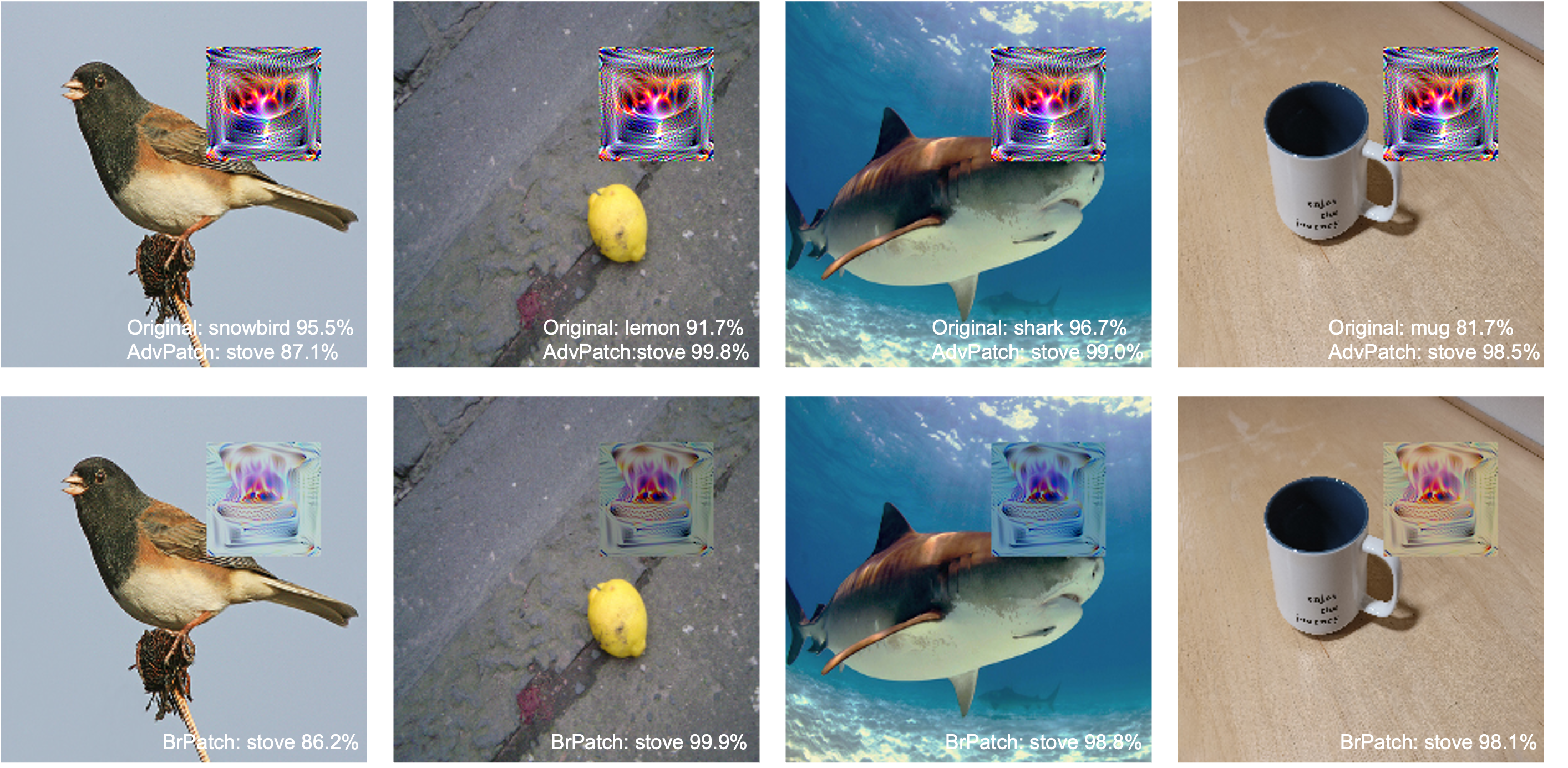}
    \caption{Images with different patches. (Top: AdvPatch; Bottom: BrPatch)}
    \label{fig:input_img}
\end{figure}

 \fref{fig:input_img} demonstrates a set of images with an original adversarial attack patch (AdvPatch) and a BrPatch (brightness range=0.24). 
 % It is observed that the BrPatch is much less visible to a human observer while keeping the same attack success rate. 
 The proposed BrPatch is observed to be less visible to a human observer in all scenarios and does not require any {additional training}. The performance of the BrPatch is also comparable to that of the AdvPatch. 
 % keeping the same attack success rate. 
\subsection{Effectiveness of different features in the physical world}
 
\subsubsection{Color transfer and texture blurring}
\begin{figure}[ht]
 \centering
 \includegraphics[width=0.85\linewidth]{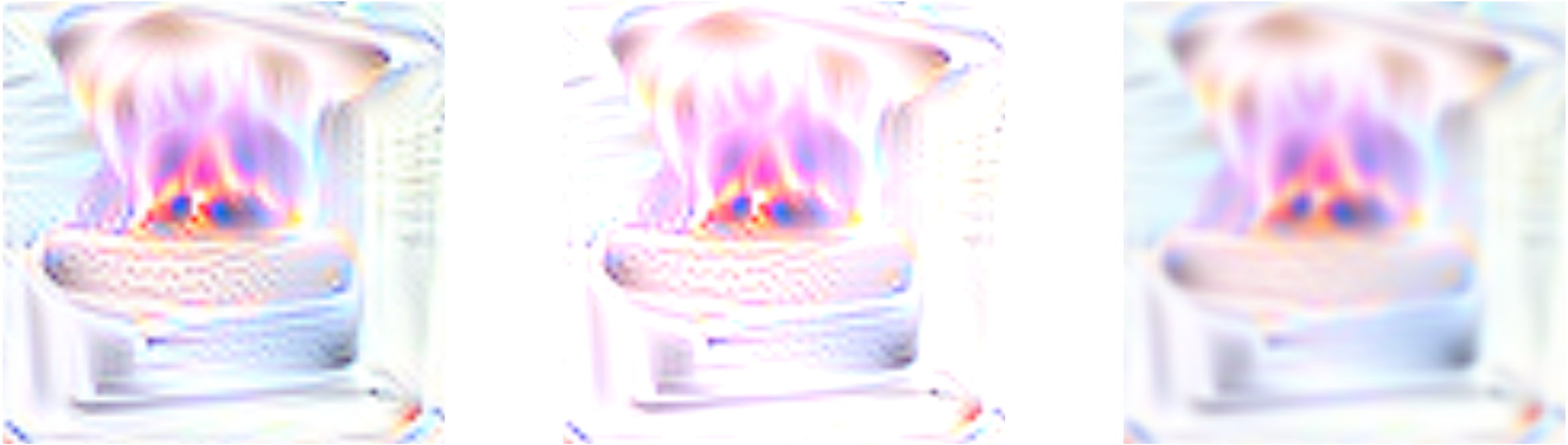}
%\centering -> This is irrelevant because of the '.5\textwidth' as Mico advised below.
    % \begin{subfigure}[t]{.25\columnwidth}
    %   \centering
    %   \includegraphics[width=.9\linewidth]{Figure/feature_adj/color_ori.png}
    %   \label{subfig:ori}
    %   \caption{Original patch}
    % \end{subfigure}
    % % \hfill
    % \begin{subfigure}[t]{.25\columnwidth}
    %   \centering
    %   \includegraphics[width=0.9\linewidth]{Figure/feature_adj/color_adj.png}
    %   \label{subfig:color}
    %   \caption{Color transfer}
    % \end{subfigure}
    % \begin{subfigure}[t]{.25\columnwidth}
    %   \centering
    %   \includegraphics[width=0.9\linewidth]{Figure/feature_adj/gaussian_adj.png}
    %   \label{subfig:filtered}
    %   \caption{Texture blurring}
    % \end{subfigure}
    \caption{Patch with different feature changes. (left: original patch; middle: color transfer patch; right: texture blurring patch)}
    \label{fig:diff_patch}
\end{figure}
 In physical-world deployments, attack patches are often affected by color transfer due to lighting conditions or blurring caused by camera focus or smudging. 
 To analyze if these changes will affect the performance of the attack patches, 
 % In order to further reduce the conspicuousness of the attack patch, we first notice the striking color of the patch. 
 % With the lightness restriction, we dramatically reduced the conspicuousness of the attack patch. However, the patch may still be visually noticeable due to its striking color and not blending into the target environment. 
 % In order to address this problem, we need to analyze if color and texture are essential to launch a successful attack. 
 % By imposing a brightness restriction on the attack patch, we have significantly diminished its conspicuousness. 
 % However, its distinctive hue may still make it perceptible to human observers. 
 % To tackle this issue, it is crucial to examine whether the color and texture of the patch are critical factors for the success of the attack. 
 % a straightforward idea is to merge the color and texture based on the background like Duan~\etal~\cite{duan2020adversarial}. 
 % Another question that arises is whether the patch needs to maintain a specific color or if it can be any color as long as it retains the texture.
 % To tackle this issue, 
 we compare the performance of an original patch, a color-adjusted patch, and a local texture adjusted patch. 
 The color-adjusted patch is applied by adding a value ($\delta$) to all values in RGB channels and making sure $\delta$ will not lead to an overflow. This color transfer will not change the texture information of the patch. 
 The local texture adjustment is applied by using a 3x3 Gaussian blur.
 % preserved the general color information of the patch while altering the local texture information.
 % A 3x3 Gaussian filter is used to apply texture filtering. This adjustment preserves the general color information while altering the local texture information.
 % original texture trend but reduces its variability, in contrast to the contrast adjustment.
 \fref{fig:diff_patch} demonstrates the two types of feature adjustments applying to a patch with brightness range=0.24.
 % lightness restrictions.

 \begin{table}[ht]
 % when using no limition one, contrast adj one also keep same acc with color one.
 \caption{Performance with color transfer and Gaussian blur}
    \label{tab:feature_adj}
    \centering
    \begin{tabular}{cccc}
    \toprule
        Brightness range  & Original   & Color transfer & Gaussian blur \\
   \midrule
        $1$ (AdvPatch)   &   89.4\%    &   90.8\%    &   47.8\%     \\
        $0.35$           &   89.5\%    &   87.9\%    &   22.7\%     \\
        $0.24$           &   74.2\%    &   75.3\%    &   10.1\%     \\
    \bottomrule
    \end{tabular}
\end{table}

 The performances of different patches are shown in \tref{tab:feature_adj}. 
 Regardless of the lightness restriction, the color transfer patch achieves almost the same success rate as the original patch. This performance shows that the patch does not need to maintain a specific color to deceive the target network. 
 On the other hand, the blurred patch exhibits a significant decrease in success rate, suggesting that local texture is the key feature in deceiving target networks. 
 % We also observe that patches with higher brightness range appear more robust to the blur distribution. 
 Using these findings, we can apply the proposed hue mapping method to adjust the color of the patch and enhance its integration with the target environment, resulting in further reduced visibility. This process does not require any learning and can be quickly applied when deploying the patch in the physical world.
 % This finding can also be applied to further decrease the visibility of BrPatch by mapping its hue to the target region.

\subsubsection{Random color variations}
 When printing an attack patch, it is important to consider that normal printers are not able to produce a patch with precisely the same color as the digital version. Therefore, 
 the patch's robustness to random color variations must be evaluated. 
 % it's essential to evaluate the patch's robustness to random color variations. 
 
 To replicate the color drift that commonly occurs during printing, we generate random noise within a restricted range that corresponds to a percentage of the original value. This approach allows us to simulate different levels of drift, and the results are shown in \tref{tab:noise}.
 % We use several different levels of random distortion on each channel 
 % We use random noise to simulate the color drift caused by a normal printer. The range of the noise is limited to a percentage of the original value to simulate different drift levels.

\begin{table}[ht]
 % ./less_10/2_70_/patch_39.pth"
    \caption{Performance with different color drift}
    \label{tab:noise}
    \centering
    \begin{tabular}{ccccc}
    \toprule
       Brightness range  &   Original    & 10\% drift  & 15\% drift   & 20\% drift\\
       \midrule
        $1$ (AdvPatch)  &    89.4\%   &  87.6\%   &  85.9\%   &   83.3\% \\
        $0.35$          &    89.5\%   &  84.2\%   &  77.6\%   &   67.2\%   \\
        $0.24$          &    74.2\%   &  68.6\%   &  43.2\%   &   27.3\% \\
        \bottomrule         
    \end{tabular}
\end{table}

\begin{figure}[ht]
    \centering
    \includegraphics[width=0.9\linewidth]{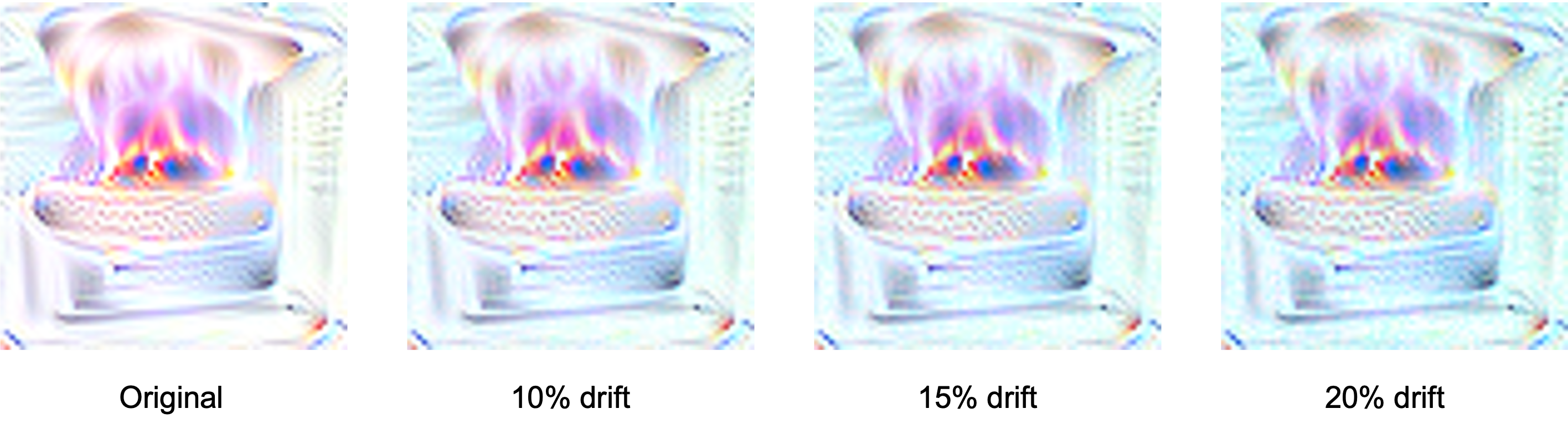}
    % \begin{subfigure}[t]{.23\linewidth}
    %   \centering
    %   \includegraphics[width=0.9\linewidth]{Figure/feature_adj/ori_adj.png}
    %   \caption{Original}
    % \end{subfigure}%
    % \begin{subfigure}[t]{.23\linewidth}
    %   \centering
    %   \includegraphics[width=0.9\linewidth]{Figure/feature_adj/01_adj.png}
    %   \caption{10\% drift}
    % \end{subfigure}
    % \begin{subfigure}[t]{.23\linewidth}
    %   \centering
    %   \includegraphics[width=0.9\linewidth]{Figure/feature_adj/015_adj.png}
    %   \caption{15\% drift}
    % \end{subfigure}
    % \begin{subfigure}[t]{.23\linewidth}
    %   \centering
    %   \includegraphics[width=0.9\linewidth]{Figure/feature_adj/02_adj.png}
    %   \caption{20\% drift}
    % \end{subfigure}
    \caption{Patches with different color drift.}
    \label{fig:noise}
\end{figure}

 We notice that even with 10\% random color drift,  the attack patches can still maintain an attack rate as reliable as that for the original clean patches.
 % maintain a reliable attack success rate like the original ones.
 % success rate doesn't significantly decrease. 
 This performance suggests that the patch is relatively robust and can withstand slight color variations during printing. 
 % Like we observe during Gaussian blur, the patches with a higher {\color{red}lightness range} appear more robust to the noise.
 \fref{fig:noise} demonstrates a set of attack patches with different noise levels. 
 As the level of noise increases, the patch experiences certain color deviations. However, adjusting the color of the patch is distinct from this process because it uniformly alters all the pixels to maintain the local texture while achieving the desired color. 
 
 % It is observed that with the noise increase, we can observe some color change in the patch. but it is different with the color adjustment, which guarantee all the pixels share the same changes (maintains the local texture)
 % with 10\% random color shifts, the success rate does not have any dramatic decrease. 
 % The feature makes it easier to deploy the patches in the physical world, which the value of each pixel may not exactly be the same value as the digital value. (due to the printer, the diffractive of the lens)
 % In physical domain deployment, the normal printer can not print a patch with exactly the same color as the digital one. So it is also necessary to analyze the robustness of the patch to random noise. 

\subsubsection{Patch scaling}
 % In previous sections, we have shown that with a bigger patch size, we can train a more powerful attack patch. However, in real-world deployment, it is very common to adjust the size of a patch to fit the environment (reduce conspicuousness / increase success rate) at the last moment. It is inefficient to retain a new patch for the new size. 
 % This lead to a question that once the patch is generated, can we boot its performance by simply using interpolation? 
 In practical deployments, it is often the case that the environment can accommodate larger sized attack patches. 
 As demonstrated in the previous sections, training larger sized patches can result in improved attack performance. 
 However, in such scenarios, there may not be enough time to train a new patch with a custom size.
 % to optimize its performance for different environments. 
 Therefore, we are motivated to investigate whether we can enhance the performance of a pre-existing attack patch by using interpolation. 
 % Specifically, we explore the potential of bilinear interpolation to expand the size of the patch and improve its success rate without requiring the creation of a new patch.
 % can be a cumbersome and time-consuming process. Therefore, we are motivated to investigate whether we can enhance the performance of a pre-existing attack patch by leveraging interpolation techniques.
 % in this section, we investigate if we can increase the performance of a small size patch by simply using interpolation.

 % In this section, we investigate the consistency of the texture,  specifically whether we can expand a given attack patch to increase its success rate. 
 We use a patch with brightness range=0.24, and its initial attack success rate is 74.2\%. Then we use bilinear interpolation to get patches with sizes 84x84, 98x98, and 112x112. The success rates of the different patches are shown in \tref{tab:diff_size}.

 \begin{table}[h]
 % 2_70_/patch_39.pth
 \caption{Performance across different sizes}
     \label{tab:diff_size}
     \centering
     \begin{tabular}{cccccc}
     \toprule
         Patch size    & 70x70 & 84x84 & 98x98 & 112x112 \\
         \midrule
         Success rate  & 74.2\% & 83.4\% & 87.4\% & 88.4\% \\
         \bottomrule  
     \end{tabular}
 \end{table}

\begin{figure}[h]
    \centering
    \includegraphics[width=0.9\linewidth]{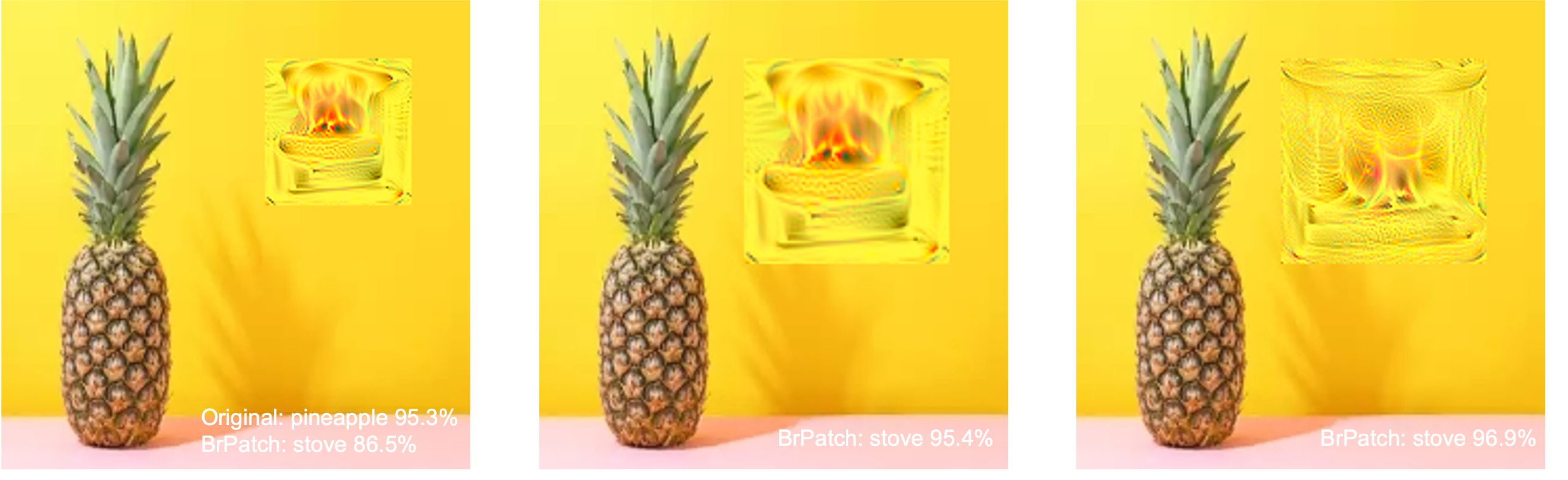}
    % \begin{subfigure}[t]{.25\columnwidth}
    %   \centering
    %   \includegraphics[width=.9\linewidth]{Figure/papple_70.png}
    %   \label{subfig:ori}
    %   \caption{}
    % \end{subfigure}
    % % \hfill
    % \begin{subfigure}[t]{.25\columnwidth}
    %   \centering
    %   \includegraphics[width=0.9\linewidth]{Figure/papple_7098.png}
    %   \label{subfig:color}
    %   \caption{}
    % \end{subfigure}
    % \begin{subfigure}[t]{.25\columnwidth}
    %   \centering
    %   \includegraphics[width=0.9\linewidth]{Figure/papple_98.png}
    %   \label{subfig:filtered}
    %   \caption{}
    % \end{subfigure}
    \caption{Patch scaling. (left: original 70x70 patch; middle: scaled 98x98 patch; right: original 98x98 patch)}
    \label{fig:patch_scale}
\end{figure}

 We notice that the attack success rate increases as the patch size grows. However, we also observe a diminishing marginal utility with each increment in size. In fact, we find that the success rate of the bilinear interpolation patch, even when it is the same size as the original patch, is lower. Some images with different scaled patches are shown in \fref{fig:patch_scale}. 

 % We first observe a corresponding increase in the success rate as the size increases. However, we also notice a diminishing marginal utility with size growth, and the success rate of the bilinear interpolation patch is lower than the original patch with the same size.
 % is also lower than the 
 % meaning that the incremental increase in success rate becomes smaller as the size gets bigger.

 % {\color{red}In order to test the robust of the patch with different feature adjustment, we need to pick a lightness restriction value which is sensitive to other changes. We use size 70, lightness restriction 0.02}\\
 % We use the attack patch (with brightness restriction) as a baseline and analyze the performance changes with color transfer, contrast adjustment, and texture filtering. 

\subsection{Physical-world attacks}
 % We further design a physical world attacking scenarios to test the \textit{BrPatch}. 
 The physical-world deployability of normal attack patches has been demonstrated in many works. However, a reasonable concern about the proposed BrPatch is whether the restricted brightness reduced the robustness of the patch in physical-world attacks. 
 In order to address this concern, we designed several physical-world deploy instances to show that the proposed attack patch can still work robustly in the physical world.
 
 We have selected two typical scenarios for the experiment: outdoor with natural light and indoor with artificial light. The artificial light has a color temperature of around 3000K (warm white), whereas the natural light has a color temperature of around 6500K (daylight). 
 For each scenario, we took several images from different angles and distances. 
 
\begin{figure}[h]
    \centering
    \includegraphics[width=0.95\linewidth]{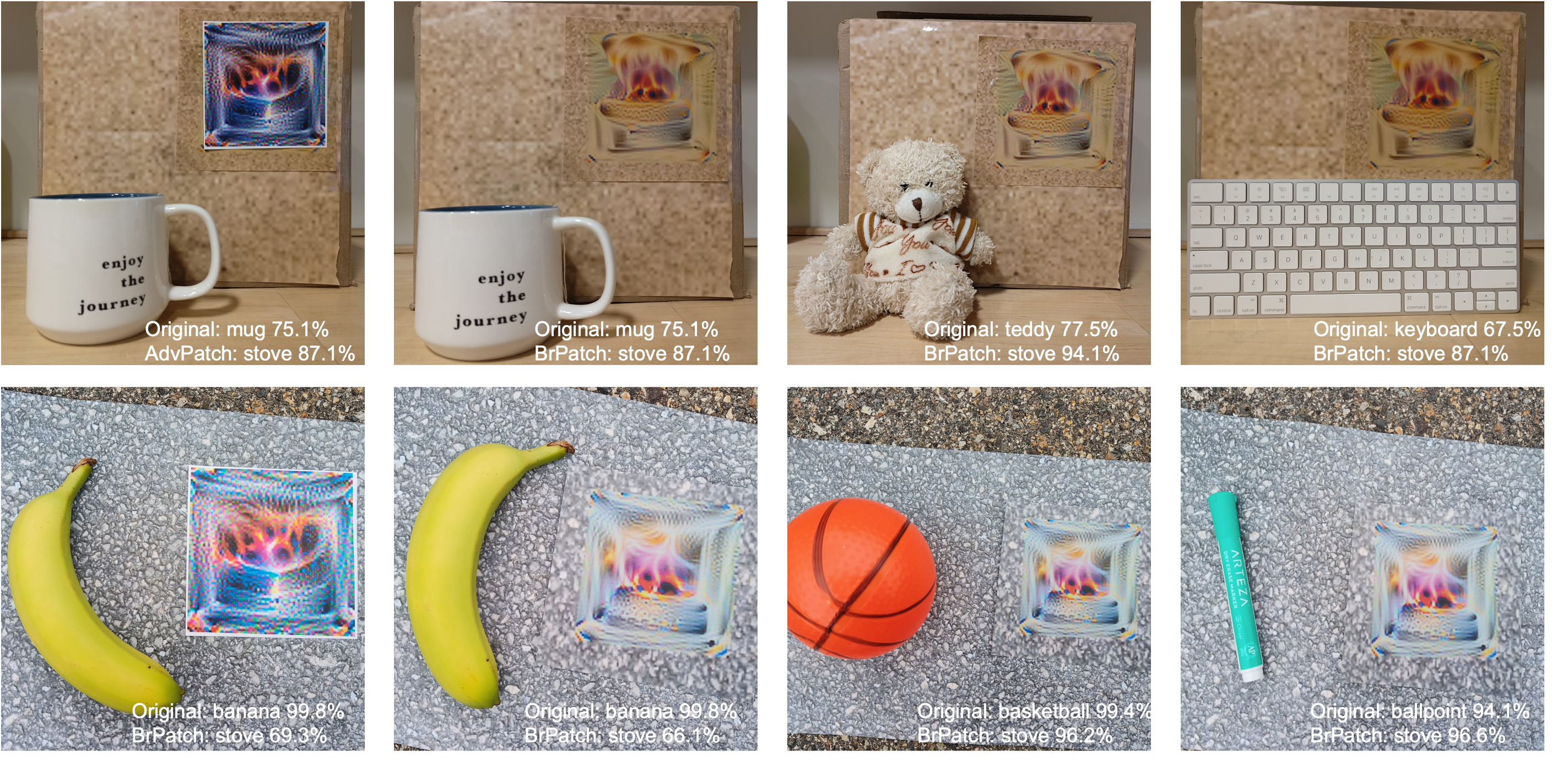}
    \caption{Comparison between AdvPatch and BrPatch in the physical world}
    \label{fig:phy_imgs}
\end{figure}

 \fref{fig:phy_imgs} shows some figures and the predictions with different patches.
 It is first observed that although the conspicuousness of the printed BrPatch is not as insignificant as in the digital domain, 
 it still greatly reduces the visibility compared to the normal attack patch (especially with the hue mapping). 
 We can then find that the probability score of the proposed patch does not obviously decrease compared to the original adversarial attack patch. 
 This experiment demonstrates that the proposed BrPatch remains sufficiently robust for physical-world deployment even when losing over 65\% of the brightness.

\section{Conclusion}

 This paper presents a novel approach to reduce the visibility of adversarial attack patches in the physical world: the brightness-restricted patch (BrPatch). 
 % Unlike existing methods that rely on specific image features or color transfers, 
 The BrPatch is the first attempt to use optical characteristics (brightness) to minimize detectability. 
 This approach allows BrPatch to maintain image independence and significantly enhances its practical applicability in physical-world attacks.
 We analyze the impact of various image features (color, texture, noise, and size) on the effectiveness of an attack patch in physical-world deployment and show that attack patches exhibit strong redundancy to brightness and are robust to color transfer and noise. 
 Our experiments also demonstrate the robustness of the BrPatch to different physical world scenarios. 
 The BrPatch presents an advanced camouflage technique, offering a reliable and effective solution for safeguarding objects or individuals from being detected by DNN-based equipment.

\bibliographystyle{ieee_fullname}
\bibliography{ijcai23}

%%%%%%%%%%%%%%%%%%%%%%%%%%%%%%%%%%%%%%%%%%%%%%%%%%%%%%%%%%%%

\end{document}